\documentclass[conference]{IEEEtran}
\IEEEoverridecommandlockouts
\usepackage{cite}
\usepackage{amsmath,amssymb,amsfonts}
\usepackage{algorithmic}
\usepackage{graphicx}
\usepackage{textcomp}
\usepackage{xcolor}
\usepackage{booktabs}
\usepackage[]{units}
\usepackage{includes}
\usepackage{multirow}
\usepackage{makecell}
\usepackage{array}
\usepackage{subfigure}

\def\BibTeX{{\rm B\kern-.05em{\sc i\kern-.025em b}\kern-.08em
    T\kern-.1667em\lower.7ex\hbox{E}\kern-.125emX}}

\usepackage{xcolor}
\usepackage[hidelinks]{hyperref}
\definecolor{darkred}{RGB}{150,0,0}
\definecolor{darkgreen}{RGB}{0,150,0}
\definecolor{darkblue}{RGB}{0,0,150}
\hypersetup{colorlinks=true, linkcolor=red, citecolor=blue, urlcolor=darkblue}

\begin{document}

\title{Multi-agent DRL-based Lane-Change Decision Model for Cooperative Platooning in Mixed Traffic\\
\author{Zeyu Mu$^{1}$, Shangtong Zhang$^{2}$ and B. Brian Park$^{3}$}
\thanks{$^{1}$Zeyu Mu is with the Link Lab and Department of Systems and Information Engineering, University of Virginia, Charlottesville, VA 22903, USA ({\tt\small dwe4dt@virginia.edu})}%
\thanks{$^{2}$Shangtong Zhang is with Department of Computer Science, University of Virginia, Charlottesville, VA 22903, USA
        ({\tt\small  shangtong@virginia.edu})}%
\thanks{$^{3}$B. Brian Park is in Department of Engineering, University of Virginia, Charlottesville, VA 22904 USA; ({\tt\small bp6v@virginia.edu})}%
}

% \author{\IEEEauthorblockN{1\textsuperscript{st} Zeyu Mu}
% \IEEEauthorblockA{\textit{Link Lab and Department of Systems and Information Engineering} \\
% \textit{University of Virginia}\\
% Charlottesville, USA \\
% dwe4dt@virginia.edu}
% \and
% \IEEEauthorblockN{2\textsuperscript{nd} Tyler Ard}
% \IEEEauthorblockA{\textit{Transportation and Power Systems (TAPS) Division} \\
% \textit{Argonne National Laboratory}
% Lemont, USA \\
% tard@anl.gov}
% \and
% \IEEEauthorblockN{3\textsuperscript{rd} Jihun Han}
% \IEEEauthorblockA{\textit{Transportation and Power Systems (TAPS) Division} \\
% \textit{Argonne National Laboratory}\\
% Lemont, USA \\
% jihun.han@anl.gov}}

\maketitle

\begin{abstract}
Connected automated vehicles (CAVs) possess the ability to communicate and coordinate with one another, enabling cooperative platooning that enhances both road capacity and traffic flow. However, during the initial stage of CAV deployment, the sparse distribution of CAVs among human-driven vehicles reduces the likelihood of forming effective cooperative platoons. To address this challenge, this study proposes a hybrid multi-agent lane-change decision model aimed at increasing CAV participation in cooperative platooning and maximizing its associated benefits. The proposed model employs the QMIX framework, integrating traffic data processed through a convolutional neural network (CNN-QMIX). This architecture addresses a critical issue in dynamic traffic scenarios by enabling CAVs to make optimal decisions irrespective of the varying number of CAVs present in mixed traffic. Additionally, a model predictive controller is designed to ensure smooth and safe lane-change execution. The proposed model is trained and evaluated within a microsimulation environment under varying CAV market penetration rates (MPRs). The results demonstrate that the proposed model efficiently manages fluctuating traffic agent numbers, significantly outperforming the baseline rule-based models. Notably, it increases cooperative platooning rates by up to 25\% over the strongest cooperative rule-based baseline while keeping lane-change activity low and, through tighter cooperative spacing, improving road capacity, showcasing its potential to optimize CAV cooperation and traffic dynamics during the early stage of deployment.
\end{abstract}


\section{Introduction}

The rapid advancement in connected automated vehicle (CAV) technologies promises significant improvements in road safety, traffic efficiency, and road capacity~\cite{Tian2018Performance}. These vehicles leverage vehicle-to-vehicle (V2V) communication, exchanging critical data such as speed, position, and acceleration, which enables CAVs to optimize decision-making processes and follow safely and closely to each other to improve the traffic capacity and efficiency. Cooperative adaptive cruise control (CACC) is one of the key technologies that enhances adaptive cruise control (ACC) by cooperating with its  connected preceding vehicle via V2V communication and then bringing benefits in terms of individual safety and comfort~\cite{CACC_bef, Aerodynamic2020Kaluva}, and road capacity and traffic efficiency~\cite{Unravelling2018Lin,Modeling2018Hao}.

Despite these advantages, the effectiveness of CACC is constrained during the early stages of CAV deployment, when low market penetration rates limit the availability of connected preceding vehicles. Under such conditions, forming and maintaining cooperative platoons becomes challenging, particularly in mixed traffic environments dominated by human-driven vehicles. Consequently, effective strategies for organizing CAVs into platoons are critical for unlocking the benefits of CACC under realistic deployment scenarios.

Prior studies have explored various approaches for facilitating cooperative platooning, including the use of dedicated CAV lanes and lane-change-based coordination strategies~\cite{Ma2019Influence,Flow2021Soomin}. While dedicated lanes can significantly increase the likelihood of platoon formation~\cite{Design2020Solmaz,Chen2024Leveraging}, their practicality is limited at low CAV penetration rates, where exclusive allocation of roadway capacity may lead to inefficiencies. In contrast, strategic lane changes that allow CAVs to follow or lead other CAVs in adjacent lanes provide a more flexible and scalable mechanism for platoon formation in mixed traffic~\cite{Lane2006Jorge, Capacity2018Danjue}. As a result, lane-change decision-making plays a central role in enabling cooperative platooning under realistic traffic conditions.

Current research on lane-change strategies for cooperative platoons often focuses on optimizing platoon formation using predefined information, such as departure times, routes, destinations, or on-demand scenarios~\cite{Liang2016Heavy,Planning2018Anirudh}. These approaches are typically designed for small-scale or homogeneous vehicle fleets and assume the availability of information that is not generally shared by individual CAVs due to privacy constraints. Alternative methods based on greedy optimization using commonly available information, such as speed and position, have been proposed~\cite{baseline}, but such approaches lack the flexibility required to handle the uncertainty, heterogeneity, and long-term interactions inherent in real-world mixed traffic.

The complexity of lane-change decision-making is further amplified in mixed traffic environments, where CAVs interact with human-driven vehicles exhibiting self-interested and often unpredictable behaviors. Human drivers typically initiate lane changes to pursue local benefits, such as higher speeds or increased spacing, which can disrupt surrounding traffic and undermine cooperative platooning. Moreover, lane changes are a major contributor to traffic safety risks, accounting for a significant proportion of traffic accidents in the United States~\cite{Hs2009AnalysisOL}. These challenges highlight the need for lane-change decision models that balance individual vehicle incentives with system-level objectives related to safety, efficiency, and cooperative behavior~\cite{Studies2017Xiang}.

Traditional lane-change decision models are predominantly rule-based, relying on handcrafted criteria to approximate human driving behavior~\cite{model1986Gipps,General2007Arne}. While these models offer interpretability, their fixed structures and manually tuned parameters limit their adaptability and generalization performance in dynamic and heterogeneous traffic environments. A second established class replaces handcrafted rules with statistical estimation, most commonly discrete-choice formulations in which the lane-change decision is modeled as a logit function of relative speeds, gaps, and driver attributes~\cite{Toledo2003Modeling}, and, more recently, supervised learning in which classifiers are trained on observed trajectories to predict whether a driver will change lanes~\cite{Hou2014Modeling}. Both classes are well suited to \emph{descriptive} tasks, that is, reproducing or forecasting the lane-change behavior of human drivers from data in which that behavior is already observed. They are less directly applicable to the \emph{prescriptive} problem addressed here, since a fitted choice model or a trained classifier imitates the maneuver distribution present in its training data and provides no mechanism for optimizing a system-level objective, such as platoon formation, that no human driver in the data was pursuing. Reinforcement learning is adopted precisely because the target behavior is not available for imitation and must instead be discovered from the long-term consequences of actions.

Recent advances in deep reinforcement learning (DRL) have demonstrated strong potential for solving sequential decision-making problems in complex and uncertain environments~\cite{Human2015Mnih,Mastering2017Silver}. In the context of autonomous driving, DRL has been widely applied to lane-change decision-making and trajectory planning, enabling vehicles to learn policies that optimize long-term performance through interaction with the environment~\cite{Wang2019Lane,Wang2022Intelligent}. Compared with rule-based methods, DRL-based approaches can better capture nonlinear interactions among vehicles and adapt to diverse traffic conditions.

To further enhance decision-making performance, recent studies have investigated improvements in observation design~\cite{Chen2019Attention, Wang2022Highway}, reward function formulation~\cite{Yuan2019Multi, Wang2022Harmonious}, and learning frameworks~\cite{Mirchevska2018High, Xu2020Reinforcement}. However, the majority of existing DRL-based lane-change models adopt a single-agent paradigm and primarily optimize ego-vehicle performance, neglecting the inherently interactive and cooperative nature of traffic systems. This limitation can lead to unsafe or inefficient behaviors when multiple CAVs operate simultaneously.

\begin{figure}[t]
\centerline{\includegraphics[width=0.8\columnwidth]{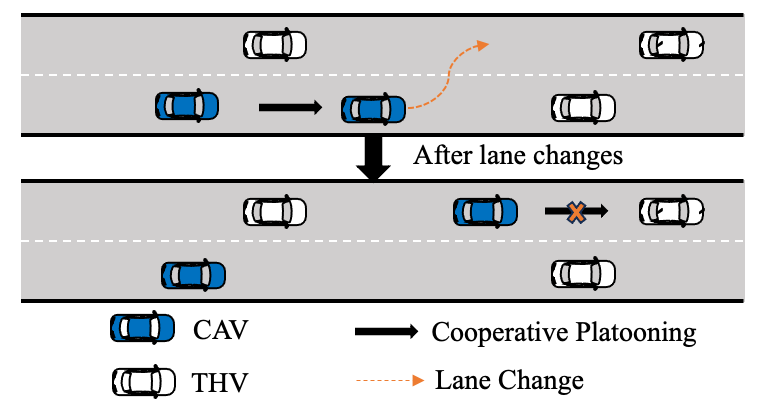}}
\caption{Lane change decision for cooperative platooning}
\label{fig:platooning}
\end{figure}

Several studies have begun to incorporate multi-agent DRL techniques to address cooperative decision-making among CAVs~\cite{Zhang2023Multi,Modeling2024Hongyu}. Despite these advances, two critical gaps remain. First, most existing multi-agent models assume a fixed number of agents during training and deployment, which is inconsistent with real-world traffic where the number of CAVs varies dynamically with market penetration rates. Second, current lane-change incentive designs often prioritize local speed or spacing gains, potentially encouraging CAVs to abandon existing cooperative platoons in favor of following faster human-driven vehicles. As illustrated in Fig.~\ref{fig:platooning}, such decisions undermine the capacity and traffic efficiency benefits of cooperative platooning and introduce additional safety risks due to the stochastic behavior of human drivers.

% \begin{figure}[t]
% \centering
%     \subfigure[Before Lane Change]{\includegraphics[width=0.6\linewidth]{Figures/beforeLC.png}}

%     \subfigure[After Lane Change]{\includegraphics[width=0.6\linewidth]{Figures/afterLC.png}}

%     \caption{Lane-Change Decision for Speed Incentive and Sabotaging Cooperative Platooning}
%         \label{fig:platooning}
% \end{figure}

To address these challenges, this study proposes a proactive cooperative lane-change strategy that enables CAVs to preserve and enhance cooperative platooning under varying market penetration rates in mixed traffic environments. A hybrid multi-agent DRL-based framework is developed, in which high-level lane-change decisions are learned using a CNN-QMIX architecture under centralized training and decentralized execution. This design allows CAVs to coordinate strategically, adapt to dynamic traffic compositions, and account for the presence of human-driven vehicles while maintaining scalable deployment. For safe lane-change implementation, the framework incorporates a model predictive control (MPC) scheme to ensure smooth and safe lane-change execution. Additionally, ACC and CACC controllers are utilized to support conventional following or cooperative platooning when necessary. 

The key contributions of this study are summarized as follows:
\begin{itemize}
    \item This study identifies the limitations of speed-oriented lane-change incentive mechanisms in existing approaches, particularly their inability to preserve cooperative platooning benefits. To address this issue, a platoon-aware lane-change strategy is proposed that prioritizes platoon formation while mitigating safety risks arising from interactions with human-driven vehicles.
    \item This work explicitly addresses the challenge of fluctuating CAV populations in dynamic traffic environments induced by varying MPRs, a critical yet underexplored factor in cooperative lane-change decision-making under realistic mixed-traffic conditions.
    \item A hybrid multi-agent DRL-based lane-change decision framework is developed to maximize cooperative platooning in mixed traffic. The framework integrates high-level decision-making using CNN-QMIX with low-level trajectory planning and control, enabling safe, scalable, and efficient execution.
\end{itemize}

This paper is organized as follows: The control framework, including the multi-agent DRL-based lane-change algorithm and lane-change planner and controller described in Section~\ref{sec:framework}. 
Section~\ref{sec:Microsimulation} gives an overview of simulation design and settings. In Section~\ref{sec:evaluation}, the training and testing results are presented and evaluated. Finally, the conclusions and future work are given in Section~\ref{sec:conclustion}.

\section{Lane-Change Decision Model} \label{sec:framework}
\subsection{Overview}
The proposed lane-change framework enables cooperative, system-level lane-change decision-making among CAVs, with the objective of facilitating platoon formation and maintenance while improving overall traffic efficiency and stability. The lane-change process is structured into two components:
i) lane-change decision-making, which determines whether and when a lane change should be initiated based on cooperative considerations, and
ii) trajectory control, which ensures safe and smooth execution of lane-change maneuvers and supports platooning behavior before and after the maneuver.

Each CAV is equipped with onboard sensing and V2V communication, acquiring the positions and speeds of nearby vehicles through sensing and of remote CAVs through communication. This information feeds a multi-agent DRL-based lane-change decision model that evaluates the \emph{collective} benefit of a lane change rather than individual gain.

When the preceding vehicle is connected, the CAV activates the CACC system to optimize its car-following behavior. In scenarios where the preceding vehicle lacks connectivity, the ACC system is utilized instead.
Based on real-time observations of surrounding traffic, the multi-agent DRL network determines whether a lane change should be initiated to improve platoon formation, gap availability, or overall traffic conditions. If a lane change is executed, the CAV updates its lane identification and recalibrates the surrounding vehicle information; otherwise, the current traffic state is maintained.

Once a lane-change decision is made, a model predictive controller (MPC) generates and tracks a smooth, dynamically feasible trajectory onto the target lane, executing the maneuver safely. This decision-making and control loop is performed iteratively at each time step until the CAV completes its trip.

\subsection{Multi-agent Lane-Change Decision Model}
In this study, the lane-change decision problem is formulated as a cooperative multi-agent reinforcement learning task, that is, as a decentralized partially observable Markov decision process (Dec-POMDP) in which each agent acts on local observations while collectively optimizing a shared objective. Let $\mathcal{A}=\{1,\dots,N\}$ denote the set of CAV agents. At time step $t$, each agent $i\in\mathcal{A}$ selects a lane-change action $a_i^t\in\mathcal{U}_i$ based on its local observation $o_i^t$. Human-driven vehicles are modeled as part of the environment dynamics and do not take learning actions.

All CAV agents share a common system-level reward $r^t$, and the objective is to learn a joint policy $\pi=\{\pi_i\}_{i\in\mathcal{A}}$ that maximizes the expected cumulative team reward,
$\max_{\pi}\; \mathbb{E}_{\pi}\!\left[\sum_{t=0}^{T} \gamma^t r^t \right]$
where $\gamma\in(0,1)$ is the discount factor. Under this formulation, the impact of a lane-change action is evaluated at the team level, as an action by a single CAV may facilitate platoon merging, stabilize traffic gaps, or improve traffic conditions for multiple neighboring or downstream CAVs.

To address the coordination and non-stationarity challenges of this cooperative setting, the QMIX algorithm is adopted. QMIX follows a centralized-training, decentralized-execution (CTDE) paradigm: agents leverage the global state during training but act on local observations at deployment, which stabilizes learning without violating decentralized execution. Its value-decomposition mechanism offers better scalability and sample efficiency than policy-based alternatives such as MADDPG~\cite{lowe2020multi}, making it well suited to the platooning-oriented, coordination-critical lane-change task considered here.

\subsubsection{QMIX}

In QMIX, each agent $i$ maintains an individual action-value function $Q_i(o_i, a_i)$, which estimates the expected cumulative reward for taking action $a_i$ given its local observation $o_i$. These individual Q-values are combined through a mixing network to form a joint action-value function $Q_{\text{total}}$, representing the system-level performance of all agents. The mixing network is a parameterized monotonic function whose weights are conditioned on the global state, so that increasing any individual $Q_i$ cannot decrease $Q_{\text{total}}$.

Concretely, the mixing network is a two-layer feed-forward network whose weights are generated by hypernetworks conditioned on the global state $s$. Stacking the per-agent utilities as $\mathbf{q} = [Q_1,\dots,Q_N]^{\top}$, the joint value is
\begin{equation}
Q_{\text{total}} = \mathbf{w}_2^{\top}\, \phi\!\big(W_1\,\mathbf{q} + \mathbf{b}_1\big) + b_2,
\label{eq:mixer}
\end{equation}
where $\phi(\cdot)=\mathrm{ELU}(\cdot)$; the first- and second-layer weights $W_1 = |h_{w_1}(s)|$ and $\mathbf{w}_2 = |h_{w_2}(s)|$ are produced elementwise by hypernetworks $h_{w_1},h_{w_2}$ and passed through an absolute value; and the biases $\mathbf{b}_1 = h_{b_1}(s)$ and $b_2 = h_{b_2}(s)$ are state-conditioned ($h_{b_2}$ being a two-layer network). The nonnegativity enforced by $|\cdot|$ yields $\partial Q_{\text{total}} / \partial Q_i \ge 0$ for every agent $i$, realizing the monotonicity constraint, while conditioning each weight on $s$ lets the mixing adapt to the global traffic state.

QMIX is trained by minimizing the temporal-difference (TD) error of the joint action-value function. The loss function is defined as
\begin{equation}
\begin{aligned}
L(\theta) =
\mathbb{E}\Big[
\big(
r + \gamma \max_{a'} Q_{\text{total}}(s', a'; \theta^{-})
- Q_{\text{total}}(s, a; \theta)
\big)^2
\Big],
\end{aligned}
\label{eq:q}
\end{equation}
where $s$ and $a$ denote the current global state and joint action, respectively, and $r$ is the shared team reward. The target network parameters $\theta^{-}$ are used to stabilize training, while the discount factor $\gamma \in (0,1)$ balances immediate and future rewards.

\subsubsection{CNN-QMIX for Varying Number of Agents} 

A key limitation of standard QMIX is its reliance on a \emph{fixed} number of agents: both the per-agent networks and the mixer are architecturally coupled to a predefined agent count, so a change in the number of agents at execution renders the learned policy incompatible. This is especially restrictive in traffic, where the number of CAVs varies as vehicles enter or leave communication range, change lanes, or exit the roadway. To overcome this while preserving the QMIX formulation, \textit{CNN-QMIX} is proposed, which replaces the fixed-size, agent-indexed input with a grid-based spatial encoding: rather than a list of per-agent features, the surrounding traffic is projected onto a discretized ego-centric map, letting the CNN extract spatial interaction features independently of the number of agents present.

\begin{figure}[t]
\centerline{\includegraphics[width=0.9\columnwidth]{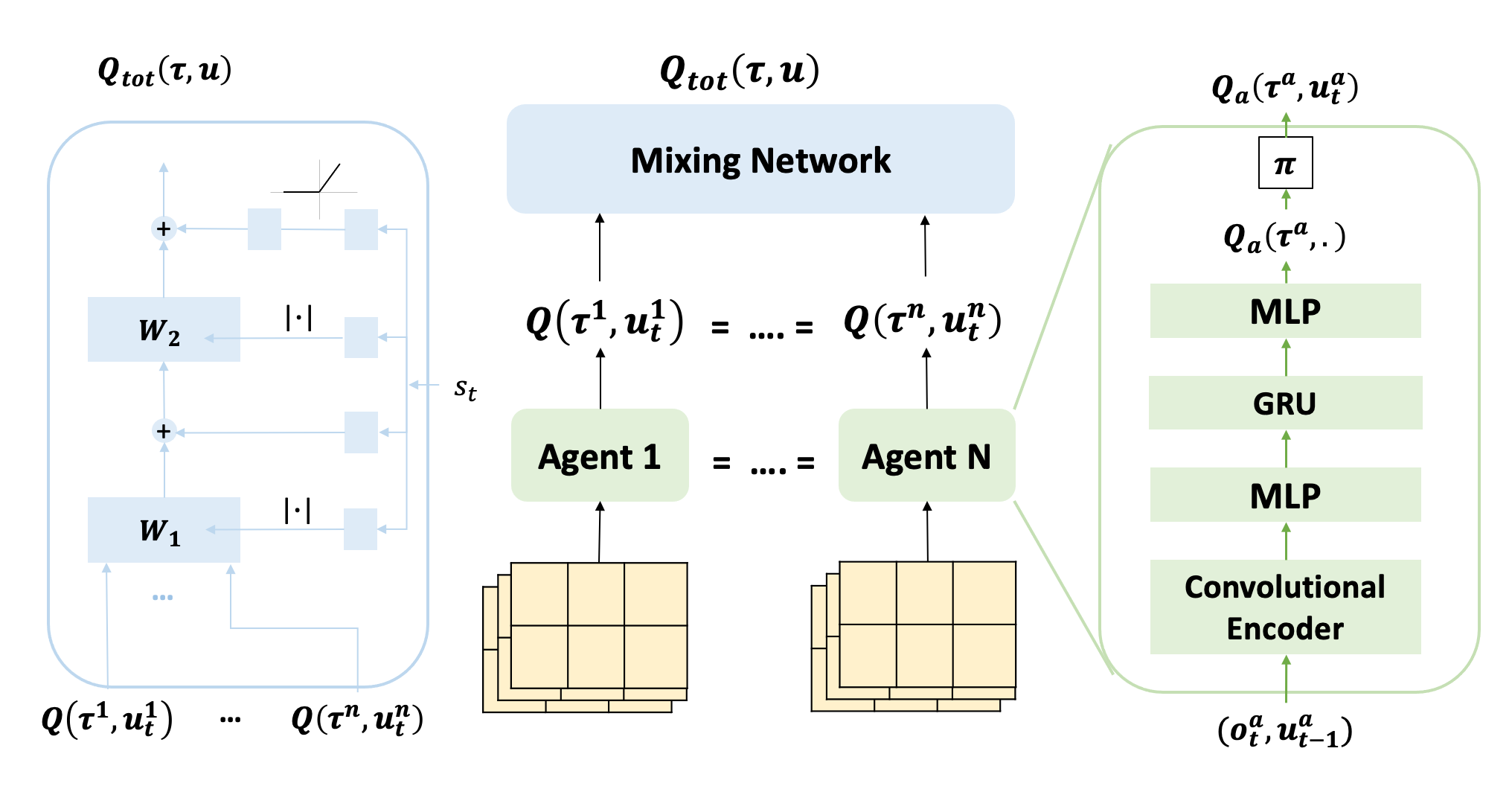}}
\caption{The Architecture of the Multi-agent DRL-based Lane-Change Decision Model}
\label{fig:RL}
\end{figure}

At time step $t$, each agent $i$ receives a local action–observation history $\tau_t^i=(o_t^i, u_{t-1}^i)$ derived from the grid-based representation, where $o_t^i$ denotes the current local observation and $u_{t-1}^i$ is the previous action. The per-agent utilities $Q_i(\tau_t^i, u_t^i)$ are mixed as in~(\ref{eq:mixer}), and action selection follows an $\epsilon$-greedy strategy.

The convolutional encoder~\cite{alzubaidi_review_2021} extracts spatial features from the grid and outputs a compact representation, which is then flattened and concatenated with one-dimensional attributes such as agent identifiers and recent actions. This combined feature vector is passed through fully connected layers with Rectified Linear Unit (ReLU) activations to produce a latent embedding.

To model temporal dependencies inherent in driving behavior, the latent features are further processed by a Gated Recurrent Unit (GRU)~\cite{Fundamentals2020Alex}, which captures historical context and temporal correlations. Finally, a fully connected output layer maps the GRU hidden state to Q-values corresponding to each admissible lane-change action. The lightweight perception backbone is intentionally adopted to emphasize the role of cooperative value decomposition and scalable decision-making in dynamic multi-agent traffic environments.

To stabilize learning, layer normalization is applied to the convolutional encoder output and to the fused spatial-kinematic feature vector prior to the GRU. Without normalization, the monotonic value decomposition is prone to a late-stage collapse in which the joint action-value estimate degrades after an initial period of stable learning. Normalizing the per-agent feature representation keeps the magnitudes of the mixed utilities well conditioned throughout training, eliminating this collapse and yielding reliable convergence over the full training horizon.

\textbf{State Space:}
Each occupied grid cell encodes the state of the corresponding surrounding vehicle relative to the ego CAV. As defined in (\ref{eq:state}), the per-cell state consists of an occupancy flag $o$, the relative longitudinal distance $\Delta x$, the relative speed $\Delta v$, the relative acceleration $\Delta a$, the vehicle type, and a lane-change intent flag $\iota$ indicating whether the occupant is currently changing lanes. The vehicle type is encoded as $\text{type} \in \{ 0, 1, 2\}$, where $0$ indicates the absence of a vehicle, $1$ represents a human-driven vehicle, and $2$ corresponds to a CAV. All continuous quantities are normalized to $[-1,1]$.
\begin{equation} s = [\,o,\ \Delta x,\ \Delta v,\ \Delta a,\ \text{type},\ \iota\,]\label{eq:state} \end{equation}

\textbf{Action Space:}
The action space consists of three discrete actions encoded as $a \in \{-1, 0, 1\}$, denoting a right lane change, keeping the current lane, and a left lane change, respectively.

\textbf{Reward}:
In the lane-change decision-making process, the reward function guides learning by encouraging decisions that lead to desirable system-level outcomes. The reward is defined at the team level and shared by all CAV agents, so that each lane-change action is evaluated by its collective effect on platoon formation, efficiency, and safety rather than by individual gain. At each decision step, the shared reward is an average over the $N$ CAV agents of five factors: traffic efficiency (speed), cooperative platoon formation and maintenance, and safety (collision avoidance and headway keeping), together with a penalty on unnecessary lane changes.

The speed term encourages CAVs to travel near the desired speed $v_d$ while saturating once it is reached, so that exceeding $v_d$ yields no additional reward:
\begin{equation}
    r_v = \frac{1}{N}\sum_{i=1}^{N}\min\!\left(\frac{v_i}{v_d},\,1\right). \label{eq:r2}
\end{equation}

Cooperative platooning is encouraged by two complementary terms. The first rewards the instantaneous fraction of CAVs traveling in a platoon,
\begin{equation}
    r_c = \frac{1}{N}\sum_{i=1}^{N}\mathbb{I}\!\left[\,i \in \text{platoon}\,\right], \label{eq:r11}
\end{equation}
where $\mathbb{I}[\cdot]$ is the indicator function. Because the instantaneous platoon rate is nearly constant and therefore only weakly sensitive to individual actions, a second \emph{sustain} term rewards \emph{maintaining} a platoon for at least $\tau_h = 3$~s,
\begin{equation}
    r_s = \frac{1}{N}\sum_{i=1}^{N}\mathbb{I}\!\left[\,h_i \ge \tau_h\,\right], \label{eq:rsustain}
\end{equation}
where $h_i$ is the time agent $i$ has continuously remained in a platoon. This makes the platooning objective action-sensitive and favors stable, long-lived platoons.

Safety is enforced by a collision term and a headway (time-to-collision, TTC) term. The collision term applies a large penalty proportional to the number of collisions $n_{\text{col}}$, while the TTC term discourages short headways to the same-lane leader:
\begin{equation}
    r_o = -\,n_{\text{col}}, \qquad
    r_t = \frac{1}{N}\sum_{i=1}^{N}
    \begin{cases}
        -1.0, & \text{TTC}_i < 1~\text{s},\\
        -0.5, & 1 \le \text{TTC}_i < 2~\text{s},\\
        \phantom{-}0, & \text{otherwise}.
    \end{cases}\label{eq:r3}
\end{equation}

Finally, a lane-change penalty discourages unnecessary maneuvers. Each executed lane change incurs a penalty of $1.0$ ($1.5$ if the CAV leaves an existing platoon), and an additional penalty is applied to changes repeated within a short time window:
\begin{equation}
    r_\ell = -\frac{1}{N}\sum_{i=1}^{N}\left(c_i + c_i^{\text{rep}}\right), \label{eq:rlc}
\end{equation}
where $c_i \in \{0,\,1.0,\,1.5\}$ encodes the per-event lane-change cost and $c_i^{\text{rep}}$ penalizes repeated changes.

The overall team reward is the weighted sum
\begin{equation}
    r = w_v r_v + w_c r_c + w_s r_s + w_o r_o + w_t r_t + w_\ell\, r_\ell, \label{eq:reward}
\end{equation}
with weights $w_v = 0.2$, $w_c = 0.3$, $w_s = 1.0$, $w_o = 5.0$, $w_t = 1.0$, and $w_\ell = 0.5$. Because the lane-change penalties are one-time events at each decision, they are scaled by the decision-interval length so that they are not diluted by the per-step reward averaging. The large collision weight ensures that safety remains the dominant consideration in the learned policy.

\begin{figure}[t]
\centerline{\includegraphics[width=0.9\columnwidth]{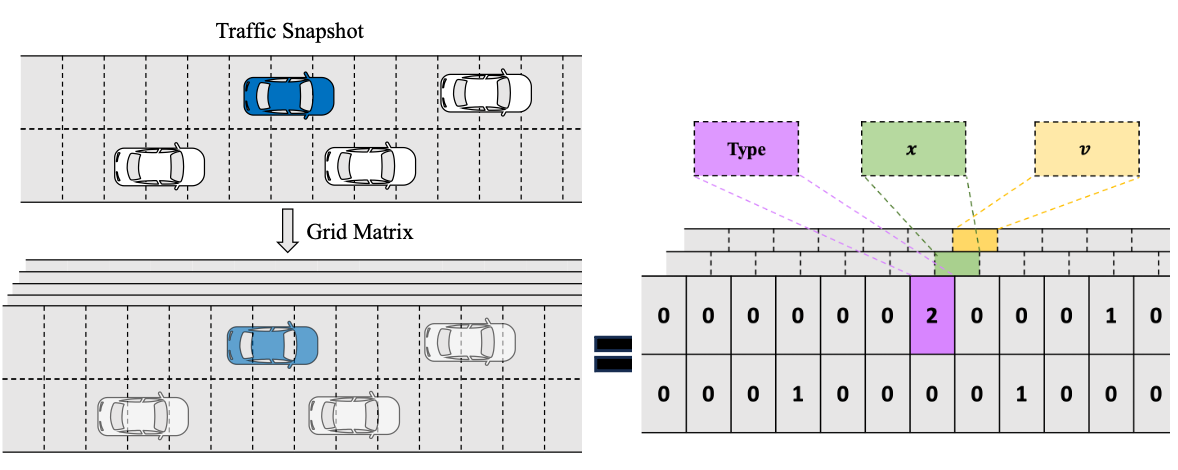}}
\caption{The Input Process for the Multi-agent DRL-based Lane-Change Decision Model}
\label{fig:input}
\end{figure}

\textbf{Input}: Each CAV's observation combines two complementary, ego-relative sources, illustrated in Fig.~\ref{fig:input}. First, a \emph{local grid} built from onboard sensing spans the three lanes centered on the ego (current and adjacent lanes) over a window from $40$~m behind to $80$~m ahead, discretized into $5$~m cells ($3 \times 24$ cells). Each cell carries the per-cell state of (\ref{eq:state}) as a separate channel, producing a $6 \times 3 \times 24$ tensor that is processed by the CNN encoder. Because the grid is ego-centric and fixed in size, it accommodates any number of surrounding vehicles.

Second, a \emph{remote V2V vector} summarizes the far field beyond the local window, up to a $300$~m communication range reflecting typical DSRC/C-V2X capabilities~\cite{Bazzi2017On}. For each of the three ego-relative lanes, a compact summary of the $80$ to $300$~m band ahead is formed from the mean relative speed, mean acceleration, number of connected vehicles, and nearest-vehicle distance, giving a $12$-dimensional vector; since HVs do not broadcast, the connected-vehicle count reflects only CAVs. The CNN feature, this remote vector, and the ego's own kinematic attributes are concatenated and fused before the recurrent layer, so the policy has both fine-grained local perception and coarse anticipatory awareness of distant platooning opportunities.

\subsection{Lane-Change Planner and Controller}
\label{sec:controller}

MARL is well suited to high-level cooperative decision-making, but its discrete lane-change commands must be rendered as smooth, dynamically feasible, and safe maneuvers. A hierarchical design is therefore adopted: the MARL policy decides \emph{whether} and \emph{which} lane change to perform, and a model predictive controller (MPC) executes it. When a change is committed, the reference is the centerline of the target lane; the MPC steers the ego CAV onto it, while the longitudinal ACC/CACC controller of Section~\ref{sec:Microsimulation} regulates speed and gap.

The vehicle follows a kinematic bicycle model with a first-order acceleration lag~\cite{Kong2015}:
\begin{subequations}
\label{eq:bicyclemodel}
\begin{align}
\dot{x} &= v\cos(\theta + \beta), &
\dot{y} &= v\sin(\theta + \beta), \\
\dot{v} &= a, &
\dot{a} &= (u_a - a)/\tau, \\
\dot{\theta} &= (v/L_r)\sin \beta, &
\beta &= \arctan\!\Big(\tfrac{L_r}{L_f+L_r}\tan u_\delta\Big),
\end{align}
\end{subequations}
where $u_a$ and $u_\delta$ are the commanded acceleration and steering, $\beta$ is the body slip angle, $\tau$ the actuator response lag, and $L_f,L_r$ the distances from the center of gravity to the front and rear axles.

For lateral control, the MPC linearizes (\ref{eq:bicyclemodel}) about the road heading. Let $e_y$ denote the lateral offset from the target-lane centerline and $e_\psi = \theta - \theta_r$ the heading error; with the small-angle slip $\beta \approx \tfrac12 u_\delta$, the lateral error dynamics reduce to
\begin{equation}
\dot e_y = v\,e_\psi + \tfrac{1}{2}v\,\delta, \qquad
\dot e_\psi = \frac{v}{L}\,\delta,
\label{eq:latmodel}
\end{equation}
with steering input $\delta \equiv u_\delta$ and wheelbase $L = L_f + L_r$. Discretizing (\ref{eq:latmodel}) with step $\Delta t$ gives $z_{k+1} = A z_k + B\,\delta_k$ for the state $z = [e_y,\,e_\psi]^{\top}$. At each control step the MPC solves the finite-horizon quadratic program
\begin{equation}
\begin{aligned}
\min_{\{\delta_k\}} \; & \sum_{k=1}^{N}\big(q_y e_{y,k}^2 + q_\psi e_{\psi,k}^2\big) \\
& \;+ \sum_{k=0}^{N-1}\big(r\,\delta_k^2 + r_\Delta(\delta_k-\delta_{k-1})^2\big) \\
\text{s.t.} \; & z_{k+1} = A z_k + B\,\delta_k, \quad |\delta_k| \le \delta_{\max},
\end{aligned}
\label{eq:mpc}
\end{equation}
and applies the first command $\delta_0$ in receding-horizon fashion. The tracking weights $(q_y, q_\psi)$ pull the CAV onto the target lane, while the effort and rate weights $(r, r_\Delta)$, scaled with $v^2$ so that the penalty acts on the lateral acceleration, yield a smooth, comfortable maneuver (about a $2.4$~s lane change with peak lateral acceleration near $2.8$~m/s$^2$). Because this cost directly shapes the closed-loop motion, the MPC jointly plans and tracks the lane-change trajectory under the steering and comfort limits, without requiring a separate polynomial reference.

\section{Microsimulation Setup} \label{sec:Microsimulation}

\begin{figure}[t]
\centerline{\includegraphics[width=0.9\columnwidth]{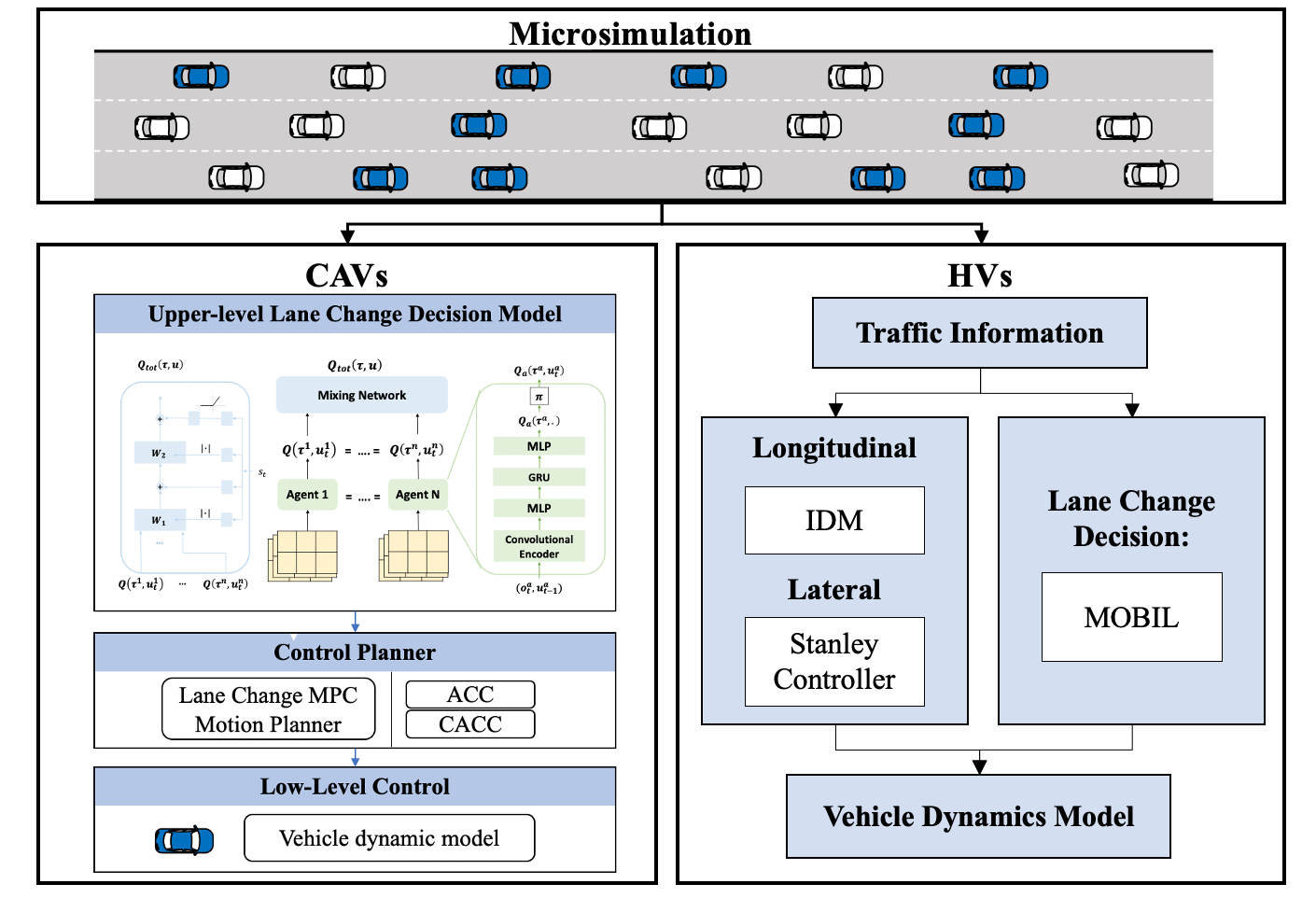}}
\caption{Microsimulation Framework}
\label{fig:microsimulation}
\end{figure}

A microsimulation framework is implemented using \texttt{Python} to model vehicle interactions and efficiently generate training data for the agents. The structure of the microsimulation is illustrated in Fig.~\ref{fig:microsimulation}. The simulator supports explicit modeling of platoon formation, join–leave logic, and cooperative lane-change constraints, which are directly coupled with the reward design and evaluation metrics in this study. To ensure safe execution, the simulator enforces two safety layers on every CAV lane change: a per-agent action-mask shield that permits a maneuver only when the time-to-collision to target-lane vehicles exceeds $2.5$~s, and a centralized concurrent-merge arbiter that prevents two CAVs from committing to the same gap on the same decision step. These layers guarantee safe overtaking independently of the reward and are applied at execution time to the learned policies and the cooperative Greedy baseline (which share the same decision pipeline); the MOBIL baseline instead relies on its own gap-acceptance safety criterion. In addition, the learned policies are equipped with a lane-change \emph{worth gate}: a commanded lane change is executed only if it is expected to be beneficial: either it places the CAV adjacent to a reachable connected vehicle in the target lane, forming or extending a platoon, or it yields a meaningful speed or gap improvement. Otherwise the maneuver is suppressed and the vehicle keeps its lane. This decision-level filter removes non-beneficial maneuvers, substantially reducing lane-change frequency and the associated traffic disturbance without degrading the platooning objective.

The MOBIL (Minimizing Overall Braking Induced by Lane Changes) model~\cite{MOBIL2007} and a cooperative rule-based method, Greedy, serve as the two rule-based lane-change baselines; human-driven vehicles use MOBIL for lane changes and the Intelligent Driver Model (IDM)~\cite{IDM2000} for longitudinal control. These components are described below. To support reproducibility, the complete simulator, training and evaluation code, and trained models are publicly available at \href{https://github.com/Zeyu2335/Cooperative-Lane-Change-Decision-Making-for-CAV-Platooning}{\textcolor{blue}{\nolinkurl{github.com/Zeyu2335/Cooperative-Lane-Change-Decision-Making-for-CAV-Platooning}}}.

\subsection{Simulator Specification and Reproducibility} \label{sec:simspec}

To make the reported results fully reproducible, the simulator is specified below in the
detail required to reconstruct any experiment in this study.

\subsubsection{Discretization and Episode Structure}
The simulator integrates vehicle dynamics at $10$~Hz, giving a physics time step of
$\Delta t = 0.1$~s. High-level lane-change decisions are issued once per second, that is,
every ten physics steps; a CAV already executing a lane change continues that maneuver
across decision ticks until it completes, and the issued high-level decision is stored
separately from the action currently being executed. Each episode runs for a fixed horizon
of $80$~s ($800$ physics steps) and terminates on that horizon rather than on a vehicle
exit condition.

\subsubsection{Road Geometry, Population, and Boundary Conditions}
The road is a three-lane segment of $3000$~m. A fixed population of $24$ vehicles is
initialized at the upstream end and driven through the segment; there is no inflow or
outflow process, and no vehicle enters or leaves during an episode, so the vehicle count
and the CAV market penetration rate are held exactly constant throughout. Vehicles are
distributed evenly across the three lanes, eight per lane. A stop sign is placed $10$~m
upstream of the downstream end of the segment as a terminal boundary condition, but at a
free-flow speed near $20$~m/s a vehicle covers at most about $1600$~m within the $80$~s
horizon, so no vehicle reaches the downstream boundary and that boundary is never
exercised in any reported experiment. This closed-population design is deliberate: it
isolates the effect of lane-change coordination on platoon formation from the confounding
effect of a fluctuating vehicle population.

\subsubsection{Initialization and Initial Speed Distributions}
For a target market penetration rate $\rho$, exactly $\lfloor N \rho \rfloor$ of the $N=24$
vehicles are drawn uniformly at random without replacement and designated CAVs; the
remainder are human-driven. Desired speeds for human-driven vehicles are drawn from a
normal distribution with mean $20$~m/s and standard deviation $2$~m/s, truncated to
$[18, 22]$~m/s, while CAVs are assigned a desired speed of exactly $20$~m/s. Every vehicle
is initialized at its own desired speed with zero initial acceleration, so no artificial
acceleration transient is introduced at the start of an episode. Within each lane, the
leading vehicle is placed uniformly in $[0, 40]$~m and each successive vehicle is placed a
gap behind its predecessor drawn uniformly from $[40, 60]$~m. Heterogeneous human driver
profiles are then applied as described in Section~\ref{sec:IDM}, assigning each
human-driven vehicle its own IDM maximum acceleration, comfortable deceleration, minimum
spacing, and desired time headway.

\subsubsection{Per-Step Update Sequence}
Each physics step proceeds in a fixed order: traffic-light states are advanced first; on a
decision tick the gap reservations held by the concurrent-merge arbiter are cleared; each
vehicle then computes its longitudinal and lateral commands in index order, ordered from
the most downstream vehicle to the most upstream, observing the current state of all other
vehicles; the dynamics of all vehicles are integrated; and collisions are finally detected
as a geometric overlap between any two vehicles. Because the arbiter accumulates gap
reservations within a single decision tick, a vehicle earlier in the loop order reserves a
contested gap first and a later vehicle requesting the same gap defers its maneuver to the
next decision tick. This ordering is therefore part of the specification of the
concurrent-merge safety layer rather than an implementation detail.

\subsubsection{Stochastic Seeding}
All pseudo-random generation derives from a single fixed seed of $20$. At the start of
each evaluation run the NumPy, Python, and PyTorch generators are seeded together, so the
sequence of traffic realizations is deterministic and byte-for-byte identical across
methods. Every controller compared in Section~\ref{sec:evaluation} therefore faces exactly
the same $100$ initial configurations and the same heterogeneous driver population at each
penetration rate, and all reported differences between methods are attributable to the
control policy rather than to sampling variation in the traffic. Learned policies are
evaluated greedily, with the exploration rate set to zero.

\subsection{Baseline Lane-Change Models}
\subsubsection{MOBIL Model for Rule-Based Lane Change}
\label{sec:mobil}

The MOBIL model is adopted as a rule-based baseline for lane-change decision-making~\cite{MOBIL2007}. For each candidate lane-change direction $\Delta\gamma \in \{1,-1,0\}$ (left, right, or keep lane), a maneuver is considered feasible only if the following incentive and safety constraints are satisfied:
\begin{subequations}
\label{eq:mobilincentive}
\begin{align}
    a' - a &> p \left(a_\mathrm{r} - a'_\mathrm{r}\right)\mathbb{I}[\gamma>1] 
            + p \left(a_\mathrm{r'} - a'_\mathrm{r'}\right) + \Delta a_\mathrm{t}, \\
    a'_\mathrm{r'} &> - b_\mathrm{s}, \\
    t - t_\ell &> \Delta t_\ell,
\end{align}
\end{subequations}
where $(\cdot)'$ denotes the predicted acceleration after a lane change.

Here, $a_\mathrm{r}$ and $a_\mathrm{r'}$ are the current and prospective rear-vehicle accelerations (from IDM), $p$ is the politeness factor, $\Delta a_\mathrm{t}$ the incentive threshold, and $b_\mathrm{s}$ the safe-braking limit; among the feasible directions, the one with the largest incentive gain is selected. When MOBIL controls the CAV baseline it uses fixed parameters ($p=0.1$, $\Delta a_\mathrm{t}=0.2~\mathrm{m/s^2}$, $b_\mathrm{s}=0.8~\mathrm{m/s^2}$), whereas human drivers use the heterogeneous MOBIL parameters of Table~\ref{tab:driver_profiles}.

\subsubsection{Greedy Platoon-Seeking Lane-Change Model}
\label{sec:cmobil}

Compared with MOBIL, which primarily emphasizes safety and politeness, the cooperative rule-based baseline, denoted Greedy, augments MOBIL with an explicit platoon-seeking objective to promote platoon formation from the perspective of each vehicle using only local information. Greedy couples the MOBIL model with a decentralized platoon-seeking rule through a conjunctive (AND) gate: a lane change is executed only if it is simultaneously endorsed by the MOBIL safety and incentive criteria and directed toward a nearby connected vehicle or platoon.

At each decision step, the ego CAV identifies the nearest connected vehicle within a search range $r$ and determines the platoon-seeking direction $\Delta\gamma_p \in \{-1, 0, +1\}$ that reduces its lateral offset to that vehicle. Independently, the MOBIL model of Section~\ref{sec:mobil} evaluates the safety and incentive of each candidate lane change and returns its preferred direction $\Delta\gamma_m$. The Greedy decision retains a maneuver only when the two agree and the target lane is safe:
\begin{equation}
\Delta\gamma =
\begin{cases}
\Delta\gamma_p, & \text{if } \Delta\gamma_p = \Delta\gamma_m \text{ and the}\\
                & \text{target lane is MOBIL-safe},\\
0, & \text{otherwise}.
\end{cases}
\label{eq:cmobil}
\end{equation}

\subsection{Longitudinal Automated Control for Cooperative Platooning}

ACC~\cite{ACC_d} is employed for longitudinal vehicle control to maintain safe inter-vehicle spacing while improving driving comfort and traffic efficiency. In this study, ACC is modeled using a proportional–derivative (PD) controller,
\begin{equation}
u(t) = k_p e(t) + k_d \dot{e}(t),
\label{eq:acc}
\end{equation}
where $k_p$ and $k_d$ are the controller gains. The spacing error $e(t)$ is defined as the difference between the desired gap $h_d(t)=v(t)T+s_0$ and the actual inter-vehicle distance, with $v(t)$ denoting vehicle speed, $T$ the desired time headway, and $s_0$ the standstill distance.

At short time headways, ACC may suffer from string instability when following unconnected vehicles~\cite{Impacts2021Mingfeng}. CACC~\cite{ACC_d} addresses this limitation by incorporating vehicle-to-vehicle (V2V) communication, enabling cooperative platooning through feed-forward compensation using the acceleration of the preceding vehicle.

Accordingly, the CACC control input augments the PD controller in (\ref{eq:acc}) with a feed-forward term,
\begin{equation}
u(t) = k_p e(t) + k_d \dot{e}(t) + f(a(t)),
\label{eq:cacc}
\end{equation}
where $a(t)$ is the acceleration of the connected preceding vehicle. The feed-forward filter $f(\cdot)$ is a first-order low-pass filter of the preceding vehicle's acceleration, designed to ensure zero steady-state spacing error following the formulation in~\cite{ACC_d}. For both ACC and CACC, the controller gains are set to $k_p=0.4$ and $k_d=0.6$, tuned to preserve string stability, with a standstill distance $s_0=4.2$~m. The desired time headway is $T=1.5$~s when following an unconnected, human-driven vehicle (ACC mode) and a shorter $T=1.0$~s when following a connected vehicle (CACC mode); the tighter CACC gap is kept string-stable by the feed-forward term and, by drawing platoon members closer, increases the road capacity analyzed in Section~\ref{sec:evaluation}.

\subsection{Human Driver Longitudinal Modeling} \label{sec:IDM}

The Intelligent Driver Model (IDM) acts as a car-following model for the native microsimulated vehicles~\cite{IDM2000}. The model commands acceleration as a function of velocity, $v$, and the relative gap and velocity difference from its preceding vehicle, $\Delta s$ and $\Delta v$
\begin{equation}\label{eq:idm}
    a_\mathrm{IDM} = a_0\left(1 - \left(\frac{v}{v_0}\right)^\delta - \left(\frac{s^*(v, \Delta v)}{\Delta s}\right)^2 \right)
\end{equation}
with the target distance function $$s^*(v, \Delta v) = s_0 + \max\left\{0, \ Tv + \frac{v\Delta v}{2\sqrt{a_0 b_0}}\right\} .$$ 

Here, $a_0$ is the maximum acceleration, $b_0$ is the comfortable deceleration, $s_0$ is the desired standstill gap, $v_0$ is the desired velocity, $\delta$ is a velocity-error exponent, and $T$ is the desired time headway.

Traffic flow in mixed environments is known to depend strongly on the composition and behavioral diversity of the driver population, so that treating human drivers as homogeneous can misstate the benefit attributed to connected vehicles~\cite{Talebpour2016Influence}. To reflect this, human-driven vehicles are modeled as a mixture of distinct driving styles rather than a single homogeneous behavior. At the start of each episode, every HV is independently assigned one of three literature-based driver profiles, conservative, normal, or aggressive, that jointly parameterize both its car-following (IDM) and lane-change (MOBIL) behavior. As summarized in Table~\ref{tab:driver_profiles}, the profiles vary the IDM maximum acceleration $a_0$, comfortable deceleration $b_0$, desired time headway $T$, and standstill gap $s_0$, together with the MOBIL politeness factor $p$, incentive threshold $\Delta a_t$, and safe braking limit $b_s$. Consequently, conservative drivers accelerate gently, maintain large headways, and change lanes rarely, whereas aggressive drivers accept tighter gaps and change lanes more readily; normal drivers lie in between. In addition, each HV is assigned a desired speed with a $\pm10\%$ variation around the nominal value under a normal distribution, and a reaction delay is introduced as a human factor with the value referenced to~\cite{mu2026formation}. This joint heterogeneity in both longitudinal and lateral behavior produces a realistic mix of driving styles within every episode and exposes the learned policy to a broad range of human behaviors during training, improving its robustness to the diversity of real human drivers.

\begin{table}[t]
\centering
\caption{Heterogeneous Human Driver Profiles (IDM Car-Following and MOBIL Lane-Change Parameters)}
\setlength{\tabcolsep}{6pt}
\begin{tabular}{lccccccc}
\toprule
& \multicolumn{4}{c}{\textit{IDM (car-following)}} & \multicolumn{3}{c}{\textit{MOBIL (lane-change)}} \\
\cmidrule(lr){2-5}\cmidrule(lr){6-8}
Profile & $a_0$ & $b_0$ & $T$ & $s_0$ & $p$ & $\Delta a_t$ & $b_s$ \\
& {\scriptsize[m/s$^2$]} & {\scriptsize[m/s$^2$]} & {\scriptsize[s]} & {\scriptsize[m]} & {\scriptsize[-]} & {\scriptsize[m/s$^2$]} & {\scriptsize[m/s$^2$]} \\
\midrule
Conservative & 0.8 & 1.5 & 2.0 & 2.5 & 1.0 & 2.0 & 2.0 \\
Normal       & 1.0 & 2.0 & 1.5 & 2.0 & 0.8 & 1.0 & 3.0 \\
Aggressive   & 2.0 & 3.0 & 1.0 & 1.5 & 0.3 & 0.3 & 5.0 \\
\bottomrule
\end{tabular}
\label{tab:driver_profiles}
\end{table}

\subsection{Simulation and Training Settings}
During training, the CAV market penetration rate is randomly sampled from $\{17\%, 25\%, 33\%\}$ at the start of each episode, exposing the single unified model to a range of penetration levels rather than a fixed one; the trained model is then evaluated at $12.5\%$, $37.5\%$, and $50\%$ to assess generalization across early-stage deployment scenarios, including the sparse regime where platoon-formation opportunities are limited.

The model parameters, network architecture, and training settings are summarized in Table~\ref{tab:model_parameters}, with the full layer-level specification available in the public code release. Hyperparameters, including the number of layers, neurons per layer, learning rate, and activation functions, are tuned through empirical testing and grid-based search, with layer depths ranging from 1 to 5 and hidden units selected from \{32, 64, 128, 256\}. The models are implemented using PyTorch and trained for 5{,}000 episodes, with model checkpoints saved every 100 episodes. During training, vehicle initial positions and human driver profiles are randomized at the start of each episode to improve generalization. To monitor generalization during training, the policy is periodically evaluated on a held-out set of validation scenarios with fixed random seeds and independently sampled heterogeneous driver profiles; the resulting validation metrics are logged alongside the training metrics and used to select the final model.

\begin{table}[t]
\centering
\caption{Model, Network, and Training Settings}
\setlength{\tabcolsep}{8pt}
\begin{tabular}{lll}
\toprule
\textbf{Symbol} & \textbf{Description} & \textbf{Value} \\
\midrule
\multicolumn{3}{l}{\textit{Human Driver Parameters}} \\
$v_0$ & Desired speed (base, $\pm10\%$) & 20 [m/s] \\
$T_d$ & Reaction delay & 0.8 [s] \\
$\delta$ & IDM acceleration exponent & 4.0 [-] \\
\multicolumn{3}{l}{\footnotesize Per-driver IDM/MOBIL params: see Table~\ref{tab:driver_profiles}} \\
\midrule
\multicolumn{3}{l}{\textit{Lane-Change and MARL Training Parameters}} \\
$L_{\text{cell}}$ & Grid cell length & 5 [m] \\
$w_v$ & Weight for speed $r_v$ & 0.2 \\
$w_c$ & Weight for platoon $r_c$ & 0.3 \\
$w_s$ & Weight for sustain $r_s$ & 1.0 \\
$w_o$ & Weight for collision $r_o$ & 5.0 \\
$w_t$ & Weight for TTC $r_t$ & 1.0 \\
$w_\ell$ & Weight for lane change $r_\ell$ & 0.5 \\
$\lambda$ & Learning rate & $1\times10^{-4}$ \\
$\Psi$ & Replay memory size & 2000 \\
$B$ & Batch size & 8 \\
$\gamma$ & Discount factor & 0.99 \\
\midrule
\multicolumn{3}{l}{\textit{Network Architecture}} \\
\multicolumn{3}{l}{\footnotesize Per-agent: CNN $3{\times}3$/32, CNN $3{\times}3$/64 (ReLU, padding 1),}\\
\multicolumn{3}{l}{\footnotesize adaptive average pooling, FC 128 and fusion FC 128 (ReLU,}\\
\multicolumn{3}{l}{\footnotesize layer normalization), GRU 128, output FC 3 ($Q$-values).}\\
\multicolumn{3}{l}{\footnotesize Mixer: state-conditioned hypernetworks, embedding 64, ELU}\\
\multicolumn{3}{l}{\footnotesize hidden 64, non-negative weights for monotonicity, scalar $Q_{\text{total}}$.}\\
\bottomrule
\end{tabular}
\label{tab:model_parameters}
\end{table}

\section{Evaluation and Analysis} \label{sec:evaluation}
\subsection{Performance Metrics}
To evaluate the performance of the proposed model, five system-level metrics are considered: platoon rate, number of lane changes, average speed, collision rate, and road capacity. The platoon rate quantifies the proportion of CAVs successfully engaged in cooperative platoons, serving as a direct indicator of coordination effectiveness. The total number of lane changes executed by all vehicles is used to assess traffic stability, as excessive lane-changing behavior can disrupt traffic flow and increase crash risk. Average speed, computed over all vehicles, is adopted as a measure of traffic efficiency, where higher values indicate smoother and more efficient flow without compromising safety. Road capacity is estimated from the realized following spacing: for each vehicle following a leader, the space headway (bumper-to-bumper gap plus vehicle length) $\bar s$ and mean speed $\bar v$ give an equivalent lane throughput $q = 3600\,\bar v / \bar s$~(veh/h/lane), so that the tighter cooperative spacing enabled by CACC platooning translates directly into higher capacity. The collision rate is reported to ensure that performance improvements are achieved without degrading safety.

\subsection{Training Evaluation}
Each model was trained for 5{,}000 episodes on a single GPU, requiring approximately ten hours of wall-clock time. 

% \begin{figure}[t]
% \centering
%     \subfigure[CNN-QMIX]{\includegraphics[width=1\linewidth]{Figures/CNN_QMIX.png}}
    
%     \subfigure[QMIX1 for MPR at 12.5\%]{\includegraphics[width=1\linewidth]{Figures/QMIX1.png}}
    
%     \subfigure[QMIX2 for MPR at 35.5\%]{\includegraphics[width=1\linewidth]{Figures/QMIX2.png}}
    
%     \subfigure[QMIX3 for MPR at 50\%]{\includegraphics[width=1\linewidth]{Figures/QMIX3.png}}
%     \caption{Training results of Multi-agent DRL-based lane change decision model, including reward and platoon rate (blue line the is moving average of 10 episodes, and orange line is moving average of 100 episodes)}
%         \label{fig:training}
% \end{figure}

\begin{figure}[t]
\centerline{\includegraphics[width=0.9\columnwidth]{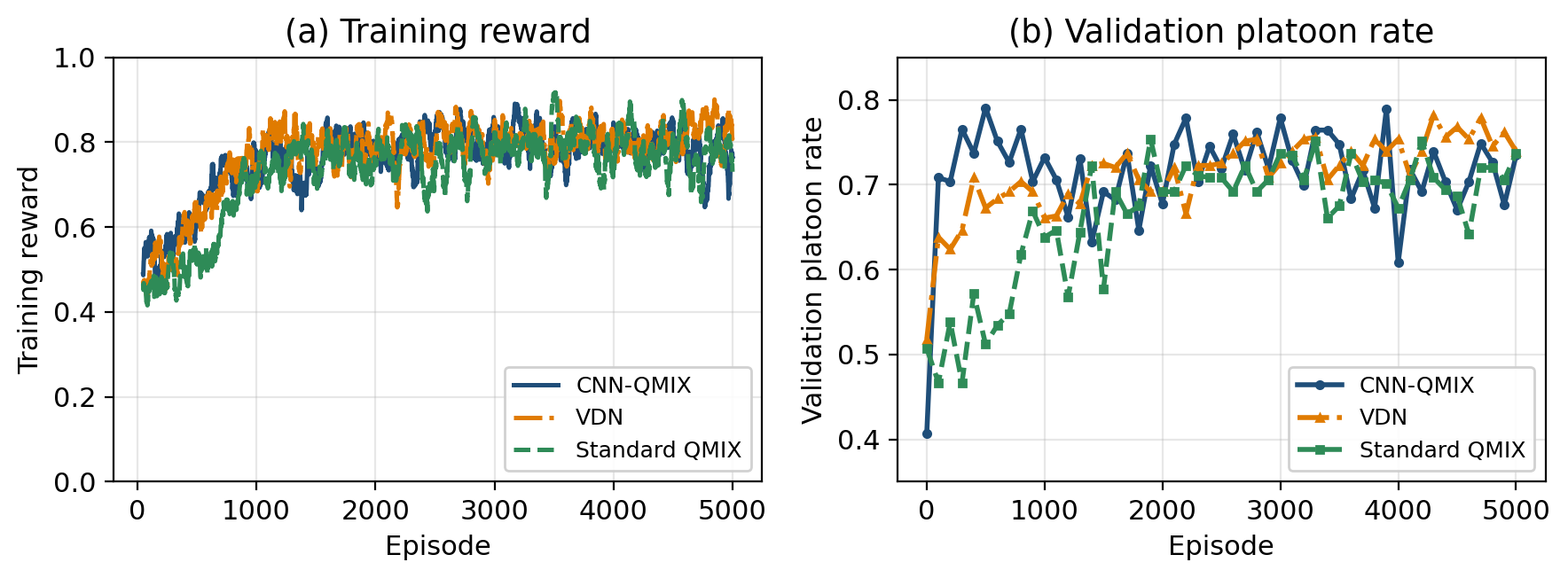}}
\caption{Training and validation results of the multi-agent DRL-based lane-change decision model, for CNN-QMIX, VDN, and the standard QMIX baseline. (a) Training reward (moving average over 50 episodes). (b) Held-out validation platoon rate.}
\label{fig:training}
\end{figure}

To disentangle the sources of performance, the proposed CNN-QMIX is compared against two learning baselines that each alter a single design choice. \emph{VDN}~\cite{VDN} keeps the CNN spatial encoding and per-agent recurrent Q-networks but replaces the state-conditioned monotonic mixer with a fixed additive decomposition, $Q_{\text{total}} = \sum_{i\in\mathcal{A}} Q_i$, isolating the contribution of the state-dependent mixing. The \emph{standard QMIX} baseline keeps the QMIX mixing network but replaces the CNN spatial grid with a fixed-size $K$-nearest-neighbor ($K=6$) MLP encoding, in which each agent observes its $K$ nearest vehicles (relative distance, speed, acceleration, and type), isolating the benefit of the spatial encoding. Both fixed-size representations still accommodate a dynamically varying number of agents; all three models are trained as single unified models across all MPRs, so the comparison isolates the effects of the value-mixing strategy and the observation representation while keeping the training procedure identical.

As shown in Fig.~\ref{fig:training}, the training and held-out validation platoon rates track each other closely without noticeable divergence, indicating stable learning under the heterogeneous-driver environment and no overfitting. Notably, CNN-QMIX attains a validation platoon rate above $0.70$ within roughly $100$ episodes, reflecting the high sample efficiency of the spatial encoding.

By comparison, both baselines converge more slowly than CNN-QMIX, and the gap widens with the severity of the design change. As shown in Fig.~\ref{fig:training}(b), VDN, which shares the CNN encoding but uses additive mixing, reaches the $0.65$ validation platoon level at roughly $400$ episodes, whereas the standard QMIX baseline, which additionally discards the spatial grid, requires roughly $900$ episodes, compared with about $100$ episodes for CNN-QMIX. Nevertheless, all three methods ultimately converge to a comparable platoon rate of about $0.74$. The primary advantage of CNN-QMIX therefore lies in sample efficiency rather than in final performance: the state-dependent monotonic mixing and, more strongly, the CNN-based spatial encoding accelerate learning while leaving asymptotic performance largely unchanged.

\begin{table}[t]
\centering
\caption{Reward weight sensitivity and ablation under $50\%$ CAV penetration ($100$ test episodes), with the worth gate disabled to isolate the effect of the reward weights.}
\setlength{\tabcolsep}{5pt}
\renewcommand{\arraystretch}{1.2}
\begin{tabular}{lccc}
\toprule
\multirow{2}{*}{Reward Setting} & Platoon & LC per & Speed \\
                                & rate    & CAV    & (m/s)  \\
\midrule
Full reward (default)           & 0.88 & 1.95 & 19.42 \\
\addlinespace[2pt]
\multicolumn{4}{l}{\textit{Platooning weight $w_c$ (default $0.3$)}}\\
\quad $w_c=0.15$                 & 0.85 & 2.04 & 19.39 \\
\quad $w_c=0.60$                 & 0.87 & \textbf{1.01} & 19.26 \\
\addlinespace[2pt]
\multicolumn{4}{l}{\textit{Lane-change penalty $w_l$ (default $0.5$)}}\\
\quad $w_l=0.25$                 & 0.84 & 2.98 & 19.47 \\
\quad $w_l=1.00$                 & 0.85 & 2.21 & 19.40 \\
\addlinespace[2pt]
\multicolumn{4}{l}{\textit{Ablations}}\\
\quad No speed ($w_v=0$)         & 0.85 & 2.61 & 19.42 \\
\quad No inst.\ platoon ($w_c=0$)& 0.84 & 2.72 & 19.43 \\
\bottomrule
\end{tabular}
\label{tab:reward_ablation}
\end{table}

To justify the reward design, Table~\ref{tab:reward_ablation} reports a sensitivity analysis in which each task-driven reward weight is varied around its default value and the policy is retrained under a $50\%$ CAV penetration rate. The safety term is held fixed throughout, since removing it admits unsafe behavior and is therefore not a meaningful operating regime. Two observations stand out. First, the platoon rate is remarkably \emph{robust}, remaining between $0.84$ and $0.88$ across every weight setting, indicating that the cooperative objective is reached through a range of incentive combinations rather than a single narrowly tuned configuration. Second, and in contrast, the lane-change frequency is highly \emph{sensitive} to the weights, ranging from $1.01$ to $2.98$ lane changes per CAV. The reward therefore genuinely shapes behavior (the environment is not under-constrained), but it does so primarily along the maneuvering-effort axis while the platooning outcome remains stable.

These trends are mechanistically consistent. Increasing the platooning weight ($w_c{=}0.60$) halves the lane-change count ($1.95\!\rightarrow\!1.01$) while holding the platoon rate, making it an attractive operating point that attains cooperation with the least disturbance. Raising the lane-change penalty $w_l$ monotonically suppresses maneuvers ($2.98\!\rightarrow\!2.21$ as $w_l$ increases from $0.25$ to $1.00$), while removing the speed term ($w_v{=}0$) leaves both the platoon rate and mean speed essentially unchanged, confirming that the speed incentive mainly trims unnecessary maneuvers rather than driving coordination. Finally, removing only the instantaneous platooning term ($w_c{=}0$) barely affects the platoon rate ($0.84$), because the sustained-platoon bonus $w_s$ carries the cooperative signal; removing \emph{both} platooning terms ($w_c{=}w_s{=}0$) reduces the platoon rate to $0.73$ and more than doubles the lane-change activity (to about $5$ per CAV), confirming that an explicit platooning incentive is essential for \emph{efficient} cooperative formation.

\subsection{Test Evaluation}

Using each trained model with the highest validation reward, 100 test simulations were conducted per method at each penetration in the same microsimulation environment, with initial vehicle positions and heterogeneous human driver profiles randomly regenerated for each test. Table~\ref{tab:results} compares all five methods, the three learning-based policies (CNN-QMIX, VDN, and the standard QMIX baseline) and the two rule-based baselines (MOBIL and Greedy), under different MPRs during the early deployment phase of CAVs.

\begin{table*}[t]
\centering
\caption{In-distribution performance across methods and market penetration rates (100 test episodes, heterogeneous drivers).}
\setlength{\tabcolsep}{9pt}
\begin{tabular}{llcccccc}
\toprule
MPR & Method & \makecell{Platoon \\ Rate} & \makecell{Max \\ Length} & \makecell{Speed \\ $[$m/s$]$} & \makecell{Capacity \\ $[$veh/h$]$} & \makecell{LC} & \makecell{Min TTC \\ $[$s$]$} \\
\midrule
\multirow{5}{*}{12.5\%} & MOBIL         & 0.10 & 1.4 & 18.5 & 1440 & 0.4 & 8.5 \\
                        & Greedy       & 0.28 & 1.5 & 18.5 & 1454 & 0.2 & 9.1 \\
                        & Standard QMIX & 0.34 & 1.7 & 18.5 & 1455 & 1.0 & 8.0 \\
                        & VDN           & 0.35 & 1.7 & 18.5 & 1456 & 0.9 & 8.5 \\
                        & CNN-QMIX      & 0.35 & 1.7 & 18.5 & 1457 & 1.2 & 8.0 \\
\midrule
\multirow{5}{*}{37.5\%} & MOBIL         & 0.52 & 3.4 & 19.1 & 1654 & 0.9 & 4.8 \\
                        & Greedy       & 0.69 & 4.0 & 19.0 & 1726 & 0.3 & 5.7 \\
                        & Standard QMIX & 0.80 & 3.7 & 19.0 & 1739 & 1.1 & 5.2 \\
                        & VDN           & 0.80 & 4.3 & 19.0 & 1739 & 1.1 & 5.0 \\
                        & CNN-QMIX      & 0.81 & 4.5 & 19.1 & 1739 & 1.3 & 5.0 \\
\midrule
\multirow{5}{*}{50\%}   & MOBIL         & 0.70 & 4.8 & 19.5 & 1809 & 1.0 & 4.3 \\
                        & Greedy       & 0.79 & 5.3 & 19.3 & 1884 & 0.4 & 5.2 \\
                        & Standard QMIX & 0.87 & 4.7 & 19.2 & 1905 & 0.5 & 4.9 \\
                        & VDN           & 0.87 & 6.1 & 19.4 & 1907 & 1.5 & 4.5 \\
                        & CNN-QMIX      & 0.89 & 5.6 & 19.4 & 1909 & 1.5 & 4.6 \\
\bottomrule
\end{tabular}
\label{tab:results}
\end{table*}

As shown in Table~\ref{tab:results}, the platoon rate of all three learning methods rises steeply with CAV penetration, from about $0.35$ at 12.5\% MPR to $0.88$ at 50\%, reflecting how much harder coordination is when CAVs are sparse. The three methods differ by at most about one percentage point at every penetration, consistent with their comparable asymptotic performance during training.

A similar pattern holds for the maximum platoon length, which grows from under two vehicles at 12.5\% MPR to roughly five to six vehicles at 50\% MPR for all three methods. Differences among the learning methods are minor: CNN-QMIX and VDN, which share the CNN spatial encoding, tend to form slightly longer platoons than the standard QMIX baseline at higher penetrations. Owing to the lane-change worth gate, all learned policies keep the number of lane changes per CAV low (at most $1.5$ at every penetration), so that platoons are formed and maintained with minimal maneuvering.

The learned policies are next compared with the two rule-based baselines of Section~\ref{sec:Microsimulation}: MOBIL, which prioritizes speed and safety and may therefore leave a platoon to pursue a higher speed, and Greedy, which adds an explicit platoon-seeking rule but remains bound by handcrafted criteria.

As shown in Table~\ref{tab:results}, all methods exhibit higher platoon rates as the MPR of CAVs rises, and CNN-QMIX attains the highest platoon rate at every penetration. Greedy ranks second, substantially outperforming MOBIL, which confirms that its explicit platoon-seeking rule captures much of the cooperative benefit that MOBIL lacks. The margin is largest where coordination is hardest: at an MPR of 12.5\%, CNN-QMIX improves the platoon rate by roughly 250\% over MOBIL and 25\% over Greedy, narrowing to about 27\% and 13\% at 50\% as platoon formation becomes easier at higher CAV densities. In terms of lane-change activity, Greedy attains its platoon rates with the fewest lane changes (0.2 to 0.4 per CAV), while the worth-gated learned policies use only modestly more (0.5 to 1.5 per CAV) to reach substantially higher platoon rates, reflecting an efficient yet more effective coordination strategy. MOBIL, despite lane-change activity comparable to or higher than Greedy, forms the fewest platoons because its maneuvers are driven by speed incentives rather than cooperation. A similar ordering holds for the maximum platoon length, where CNN-QMIX consistently forms among the longest platoons.

\begin{figure}[t]
\centerline{\includegraphics[width=0.9\columnwidth]{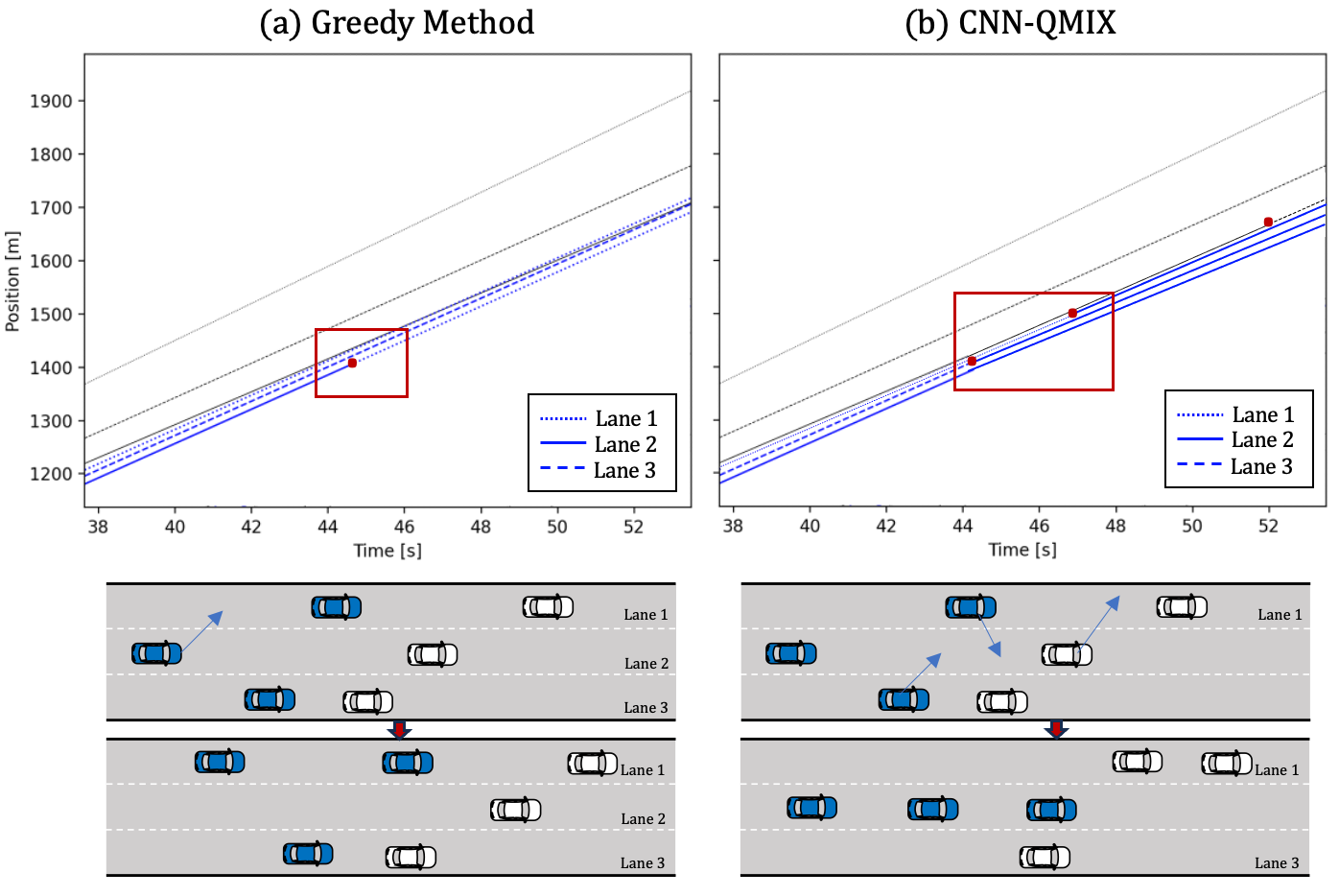}}
\caption{Illustrative Case Study: Time–Space Diagrams and Traffic Snapshots Comparing Greedy and CNN-QMIX Lane-Change Strategies}
\label{fig:example}
\end{figure}

Fig.~\ref{fig:example} presents an illustrative case study comparing the Greedy strategy and the proposed CNN-QMIX approach. In this scenario, three CAVs are initially located in three different lanes. The CAV in Lane~1 follows a higher-speed HV, whereas the CAVs in Lanes~2 and~3 are constrained by slower HVs. Under the Greedy strategy, each CAV independently seeks the nearest connected vehicle to form a platoon using only local information. As a result, the CAV in Lane~2 performs a lane change to follow the CAV in Lane~1 in order to form a platoon and benefit from the higher leading speed. However, this myopic decision leaves the CAV in Lane~3 isolated, preventing the formation of a fully cooperative CAV platoon.

In contrast, the CNN-QMIX strategy enables coordinated and system-level decision-making among CAVs. Specifically, the CAVs in Lanes~1 and~3 jointly execute lane-change maneuvers to form a three-vehicle cooperative platoon. Meanwhile, the lane change of the CAV in Lane~1 creates sufficient space for the HV originally in Lane~2 to merge into Lane~1 and follow a higher-speed vehicle. Consequently, all three CAVs reach their desired speeds while maintaining a stable platoon, which the myopic Greedy strategy cannot achieve.

The efficiency metrics in Table~\ref{tab:results} show that all methods achieve similar mean speeds (about $18.5$ to $19.4$~m/s, averaged over all vehicles), with the speed-seeking MOBIL baseline marginally the fastest and the standard QMIX baseline marginally the slowest. Road capacity, in contrast, increases substantially with CAV penetration, from about $1450$ to $1900$~veh/h/lane between $12.5\%$ and $50\%$ MPR (a ${\sim}30\%$ gain), as more vehicles form tightly spaced CACC platoons. At each penetration the learning-based methods attain the highest capacity: at $50\%$ MPR they reach about $1905$~veh/h/lane, versus $1884$ for Greedy and $1809$ for MOBIL, because their higher platoon rates translate the tighter CACC following gap into denser flow. Consequently, CNN-QMIX combines the highest platoon rate and longest platoons with the greatest road capacity, at only a marginal cost in average speed; its principal advantages are strong cooperative platoon formation and fast, sample-efficient convergence.

To assess whether the learned policy maintains safe lane-change behavior, two safety indicators are examined over 100 test episodes: the per-episode collision rate and the minimum time-to-collision (TTC) to the same-lane leader, the latter reported in Table~\ref{tab:results}. All three learning-based methods (CNN-QMIX, VDN, and the standard QMIX baseline) achieve a collision rate of \emph{zero} across every penetration, and the minimum TTC remains above $4$~s, far exceeding the ${\sim}1.5$~s critical value commonly associated with rear-end conflict. The full TTC distribution is strongly skewed toward safe values: fewer than $1.5\%$ of closing samples fall below $5$~s and none fall below $2$~s for any method, so same-lane (rear-end) safety is consistently preserved. The small residual collision rate of the rule-based baselines (up to $2\%$ for MOBIL, whose maneuvers are driven purely by speed incentives) arises from lateral lane-change/merge conflicts rather than rear-end closing. This safety robustness reflects the layered design of the framework: safe headways are maintained by the CACC/ACC longitudinal controller; each candidate lane change must pass an action-mask safety check (a minimum $2.5$~s TTC to target-lane vehicles); a centralized concurrent-merge arbiter prevents two CAVs from committing to the same gap on the same decision step; and the reward's collision and TTC penalties provide additional shaping. 

\subsection{Robustness to Human-Model Distribution Shift}
To assess whether the learned coordination generalizes beyond the human behaviors seen during training, CNN-QMIX is evaluated under an out-of-distribution (OOD) perturbation of the entire human driver population. At test time, every heterogeneous human's car-following (IDM $a$, $b$, $T$, $s_0$) and lane-change (MOBIL) parameters are scaled by a fixed factor of $\pm 5\%$ and $\pm 10\%$, with each driver retaining its shifted parameters deterministically for the whole episode. A positive shift makes the population more aggressive (shorter headways and gaps, higher accelerations, and lower politeness and lane-change thresholds), whereas a negative shift makes it calmer; in both cases the aggregate human behavior moves off the training distribution. The policy weights are frozen, and $100$ episodes are evaluated per shift at a representative $50\%$ MPR.

\begin{figure}[t]
\centerline{\includegraphics[width=\columnwidth]{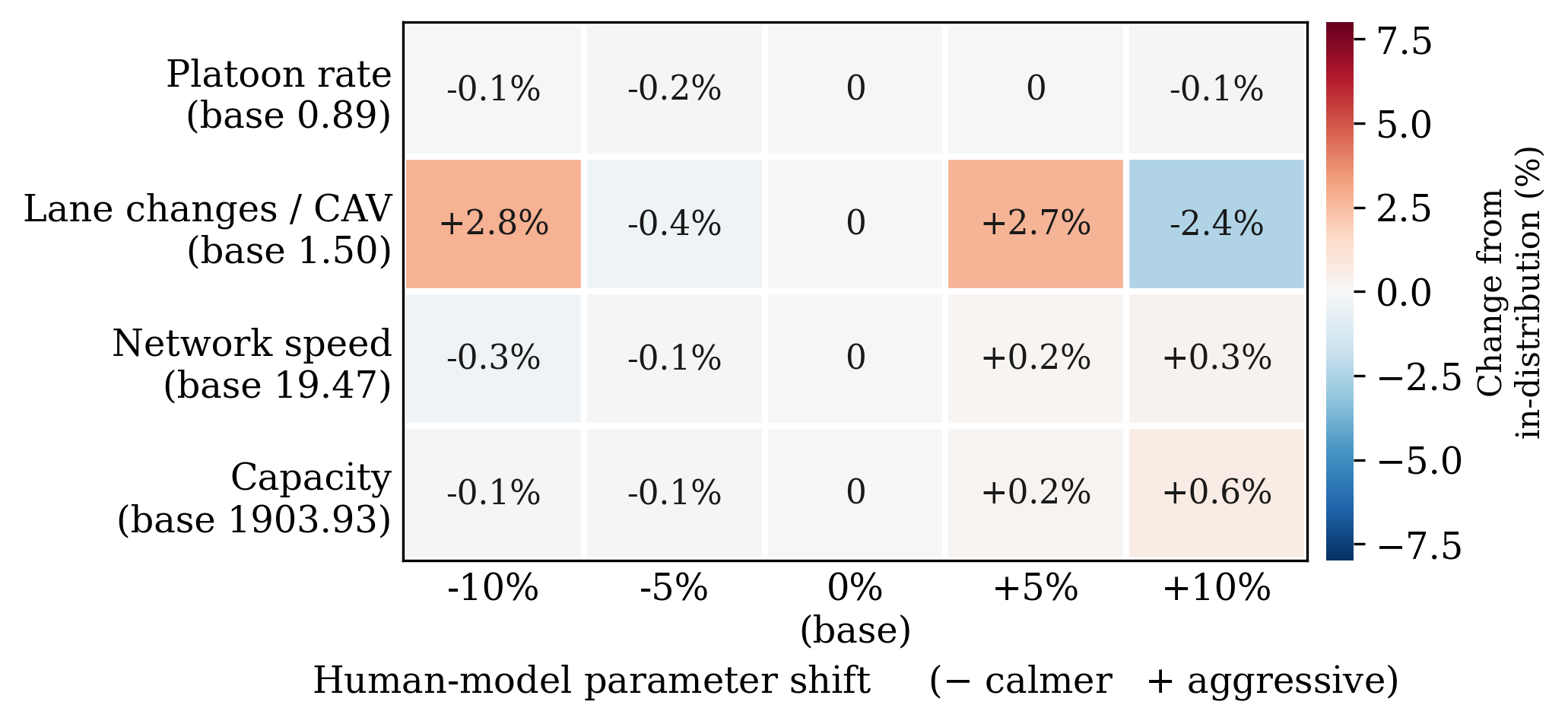}}
\caption{Out-of-distribution robustness of CNN-QMIX at $50\%$ MPR ($100$-episode). Each heterogeneous human's car-following and lane-change parameters are shifted by $\pm 5\%$ and $\pm 10\%$ (positive $=$ more aggressive, negative $=$ calmer). Each cell reports a metric's percentage change from its in-distribution value (bases given in the row labels); pale cells (near zero) indicate insensitivity, and the $0\%$ column is the in-distribution reference.}
\label{fig:ood}
\end{figure}

Fig.~\ref{fig:ood} reports each metric's percentage change from its in-distribution value as a heatmap, and shows that the policy is essentially insensitive to this shift. The platoon rate, network speed, and road capacity all deviate by well under $5\%$ across the full $\pm 10\%$ range, and even the most sensitive metric, lane-change frequency, changes by at most about $4\%$. These residual deviations are non-monotonic and lie within the sampling noise of the $100$-episode evaluation (for example, the platoon-rate standard error is about $0.7\%$), so the shift induces no systematic change in behavior. This robustness is consistent with the ego-centric CNN representation, which encodes the \emph{relative} kinematics of surrounding vehicles rather than the absolute parameters of any particular human model; a uniform change in the driving population therefore leaves the decision-relevant features largely unchanged. The result indicates that the cooperative platooning benefit is not an artifact of the specific human calibration used in training and is expected to persist under realistic variation in human driving styles.

\section{Conclusions and Future Work} \label{sec:conclustion}

This study proposed a multi-agent lane-change strategy for cooperative platooning built on a high-level multi-agent deep reinforcement learning (DRL) decision model that accommodates a dynamically varying number of agents. By encoding surrounding traffic as a spatial grid processed by a CNN, a single unified model operates across market penetration rates without MPR-specific retraining. Against a neighbor-list standard QMIX baseline and a VDN baseline trained under an identical protocol, the proposed CNN-QMIX learns substantially faster and more stably while reaching comparable final performance, so its advantage lies in sample efficiency rather than in asymptotic quality. A team-level reward that combines speed and safety with explicit platooning incentives, together with a lane-change worth gate that suppresses non-beneficial maneuvers, yields higher platoon rates than the MOBIL and Greedy rule-based baselines while keeping lane-change activity low and improving road capacity through tighter cooperative spacing. At execution, an MPC controller tracks the intended trajectory under steering and comfort limits, coupling strategic decision-making to dynamically feasible motion.

Despite the encouraging results, this study has several limitations that motivate future research. First, the proposed framework is currently evaluated in highway scenarios and requires extension to more complex environments, such as urban arterials with signalized intersections. Addressing these settings will require refinements to both the reward formulation and model structure to ensure robustness under diverse and highly interactive traffic conditions. Second, key model parameters, including reward weights, are manually specified in the current implementation. Future work will investigate treating these parameters as learnable components, enabling automatic adaptation through training and reducing reliance on manual tuning. Third, although the simulator is documented in full and released as open source to support reproducibility, cross-validation of its emergent traffic dynamics against an established platform such as SUMO is a valuable further step and is left to future work. Finally, an important research direction is to systematically examine the trade-off between architectural complexity and real-time deployability, particularly for large-scale simulations and field implementations. Addressing these challenges will further enhance the practicality and scalability of DRL-based cooperative control strategies for automated vehicular systems.

\bibliographystyle{./IEEEtran}
\bibliography{./IEEEabrv,./IEEEexample}

\vspace{-40pt}

\begin{IEEEbiography}
[{\includegraphics[width=1in,height=1.25in,clip,keepaspectratio]{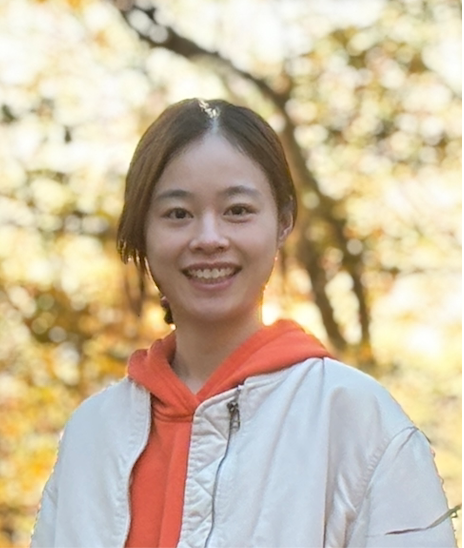}}]{Zeyu Mu} received her M.S. degree in Electrical Engineering from the University of Southern California in 2020. She is currently a Ph.D. candidate with the Link Lab and the Departments of Systems \& Information Engineering at the University of Virginia. Her research interests include control, optimization, and intelligent decision-making in connected and automated vehicles.
\end{IEEEbiography}

\vspace{-50pt}
\begin{IEEEbiography}[{\includegraphics[width=1in,height=1.25in,clip,keepaspectratio]{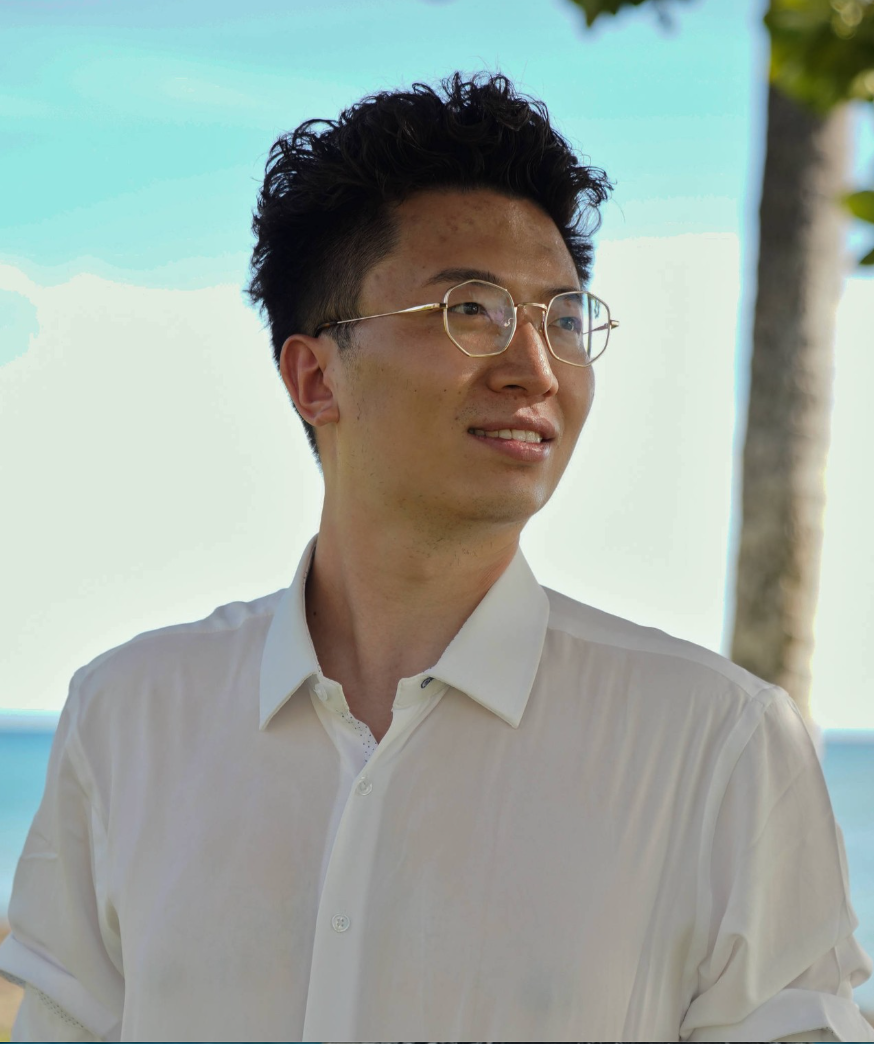}}]{Shangtong Zhang} is an Assistant Professor with the Department of Computer Science at the University of Virginia. He has published over 40 journal articles and conference papers on reinforcement learning and artificial intelligence. His research focuses on designing theoretically grounded reinforcement learning algorithms that have practical impacts.

\end{IEEEbiography}

\vspace{-50pt}

\begin{IEEEbiography}[{\includegraphics[width=1in,height=1.25in,clip,keepaspectratio]{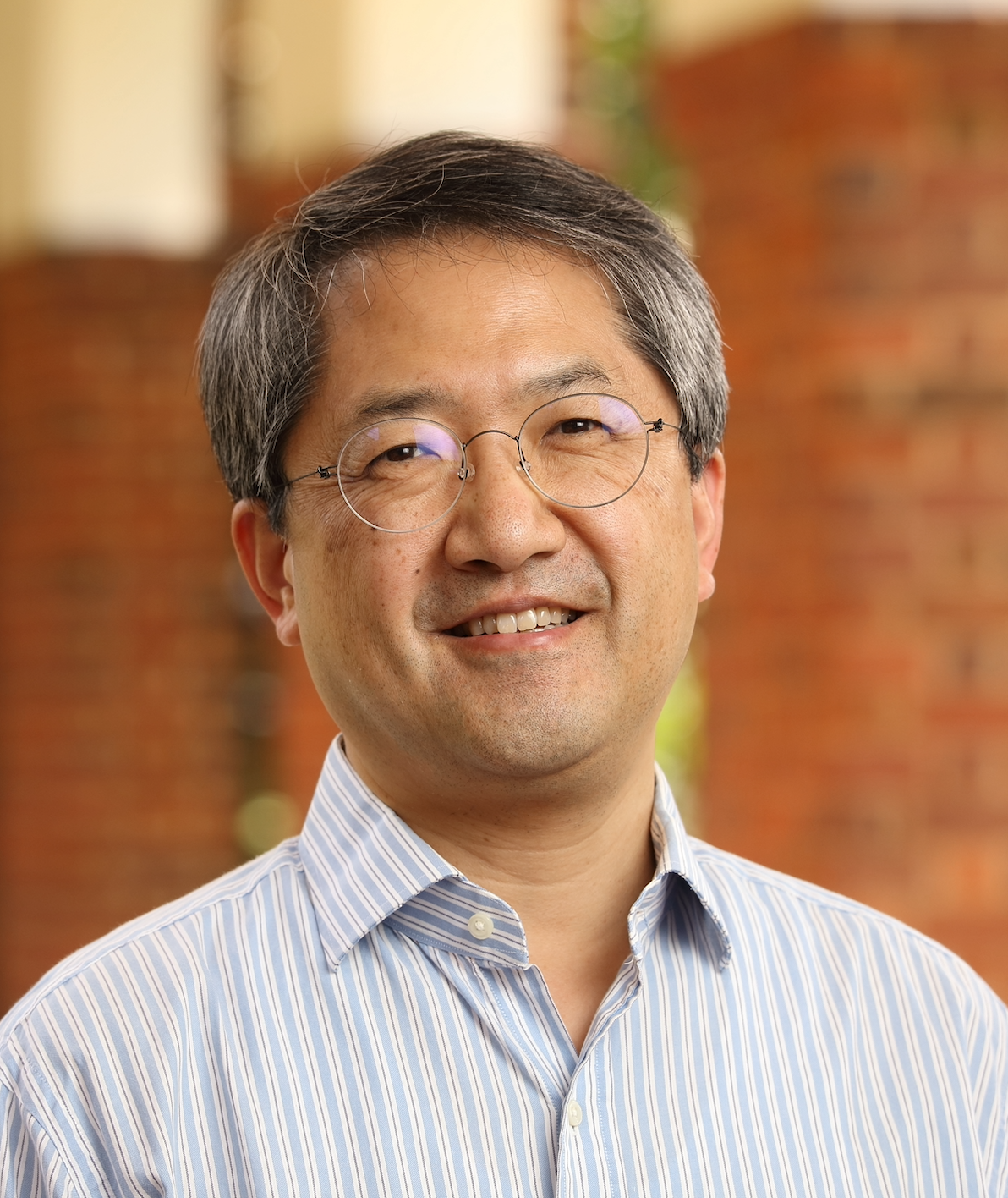}}]{Byungkyu Brian Park} (Senior Member, IEEE) is a Professor with the Link Lab and the Departments of Civil \& Environmental Engineering and Systems \& Information Engineering at the University of Virginia. He has published over 180 journal articles and conference papers on transportation system operations, management, and intelligent transportation systems. His research interests include cyber-physical systems for transportation, stochastic optimization, and connected and automated vehicles.
\end{IEEEbiography}

\end{document}